\documentclass[runningheads]{llncs}
\usepackage[utf8]{inputenc}

\usepackage[english]{babel}

\usepackage[style=lncs, 
			backend=biber, 
			isbn=false, 
			doi=false, 
			maxbibnames=10,
			useprefix=true,
			firstinits=true,
			terseinits=true,
			]{biblatex} 
\usepackage{subfig}

\usepackage{textcomp}

\AtEveryBibitem{
	\clearfield{isbn}
	\clearfield{doi}
	\clearfield{publisher}
	\clearfield{pages}
	\ifentrytype{inproceedings}{
			\clearfield{url} 
			\clearname{editor}
		}{
	}
	\ifentrytype{book}{
		\clearfield{url}
		}{
	}
	\ifentrytype{article}{
			\clearfield{url}
		}{
	}
	\ifentrytype{inbook}{
			\clearfield{url}
		}{
	}
}
\DeclareDataInheritance{proceedings}{inproceedings}{
	\noinherit{editor}
	\noinherit{series}
	\noinherit{volume}
	\noinherit{publisher}
}
\DeclareDataInheritance{book}{inbook}{
	\noinherit{editor}
	\noinherit{series}
	\noinherit{volume}
}

\usepackage{booktabs}
\usepackage{paralist}

\usepackage{tikz}
\usepgfmodule{shapes}
\usetikzlibrary{automata,petri,arrows}
\usetikzlibrary{positioning}

\usepackage{pgfplotstable}
\usepackage{pgfplots}

\newcommand{\todo}[1]{}
\newcommand{\todotitle}[1]{}
\newcommand{\todonote}[1]{}

\usepackage{csquotes}

\usepackage{xspace}

\newcommand{\ie}{i.\,e.\@\xspace}
\newcommand{\eg}{e.\,g.\@\xspace}
\newcommand{\cf}{cf.\@\xspace}

\newcommand{\mathtextit}[1]{\text{\textit{#1}}}
\newcommand{\mathtexttt}[1]{\text{\texttt{#1}}}
\newcommand{\mathtextsc}[1]{\text{\textsc{#1}}}

\usepackage{amsmath} 

\usepackage{etoolbox}
\apptocmd\normalsize{
 \belowdisplayskip     10.0pt plus 0.0pt minus 5.0pt
}{}{}

\usepackage[
pdfusetitle
]{hyperref} 
\hypersetup{
    colorlinks=true,
    linkcolor=black,
    citecolor=black,
    filecolor=black,
    urlcolor=black
}

\makeatletter
\g@addto@macro\UrlSpecials{\camelurl}
\def\camelurl{%
\count@`a
\loop
\mathcode\count@"8000
\uccode`\~\count@\uppercase{\edef~{\mathchar\the\count@\noexpand\breakifupper}}%
\ifnum\count@<`\z
\advance\count@\@ne
\repeat}

\def\breakifupper#1{%
\ifcat .\noexpand#1%
\ifnum`#1>40
\ifnum`#1<91
\penalty\z@
\fi\fi\fi
#1%
}

\makeatother

\title{Specifying, Monitoring, and Executing Workflows in Linked Data Environments}
\titlerunning{Specifying, Monitoring, and Executing Workflows in Linked Data}

\let\oldand\and
\renewcommand\and{\texorpdfstring{\oldand}{and\xspace}}

\author{Tobias Käfer \and Andreas Harth}

\institute{Institute AIFB, Karlsruhe Institute of Technology (KIT), Germany\\\email{\{tobias.kaefer|harth\}@kit.edu}}

\begin{document}

\maketitle

\begin{abstract}
	
We present an ontology for representing workflows over components with Read-Write Linked Data interfaces and give an operational semantics to the ontology via a rule language.
Workflow languages have been successfully applied for modelling behaviour in enterprise information systems, in which the data is often managed in a relational database.
Linked Data interfaces have been widely deployed on the web to support data integration in very diverse domains, increasingly also in scenarios involving the Internet of Things, in which application behaviour is often specified using imperative programming languages.
With our work we aim to combine workflow languages, which allow for the high-level specification of application behaviour by non-expert users, with Linked Data, which allows for decentralised data publication and integrated data access.
We show that our ontology is expressive enough to cover the basic workflow patterns and demonstrate the applicability of our approach with a prototype system that observes pilots carrying out tasks in a mixed-reality aircraft cockpit.
On a synthetic benchmark from the building automation domain, the runtime scales linearly with the size of the number of Internet of Things devices.

\end{abstract}

\section{Introduction}

Information systems are increasingly distributed.
Consider the growing deployment of sensors and actuators, the modularisation of monolithic software into microservices, and the movement  to decentralise
data from company-owned data silos into user-owned data pods.
The drivers of increasing distribution 
include:

\begin{itemize}
\item Cheaper, smaller, and more energy-efficient networked hardware makes wide\-spread deployment feasible\footnote{\url{http://www.forbes.com/sites/oreillymedia/2015/06/07/how-the-new-hardware-movement-is-even-bigger-than-the-iot/}} 
\item Rapidly changing business environments require flexible re-use of components in new business offerings~\cite{DBLP:books/lib/Newman15}
\item Fast development cycles require independent evolution of components~\cite{DBLP:books/lib/Newman15}
\item Privacy-aware users demand to retain ownership of their data\footnote{\enquote{Putting Data back into the Hands of Owners}, \url{http://tcrn.ch/2i8h7gp}} 
\end{itemize}
The distribution into components raises the opportunity to create new integrated applications out of the components, given sufficient interoperability.
One way to make components interoperable is to equip the components with uniform interfaces using technologies around Linked Data.
Consider, \eg, the W3C's Web of Things\footnote{\label{foot:wot}\url{http://www.w3.org/WoT/}} initiative and the MIT's Solid (``social linked data'') project\footnote{\url{https://solid.mit.edu/}}, where REST provides an uniform interface to access and manipulate the state of components, while RDF provides an uniform data model for representing component state that allows using reasoning to resolve schema heterogeneity.
While the paradigms for (read-only) data integration systems based on Linked Data are relatively agreed upon~\cite{DBLP:books/crc/linked2014}, techniques for the creation of applications that integrate components with Read-Write Linked Data interfaces are an active area of research~\cite{DBLP:conf/rest/VerborghSDCVW12,DBLP:conf/atal/CiorteaBZF17,DBLP:conf/esws/CapadisliG0ASB17,DBLP:conf/www/KaferH18}.
Workflows are a way to create applications, according to \citeauthor{DBLP:books/daglib/0090620}~\cite{DBLP:books/daglib/0090620}, that is highly suitable for integration scenarios, easy to understand (for validation and specification by humans), and formal (for execution and verification by machines).
E.\,g., consider an evacuation support workflow for a smart building (\cf task 4 in our evaluation, Section~\ref{sec:evaluation}), which integrates multiple systems of the building, should be validated by the building management and the fire brigade, verified to be deadlock-free, and executable.
Hence, we tackle the research question:
\emph{How to specify, monitor, and execute applications given as workflows in the environment of Read-Write Linked Data?}

\todonote{Why interesting?}

The playing field for applications in the context of Read-Write Linked Data is big and diverse:
As of today, the Linking Open Data cloud diagram\footnote{\url{http://lod-cloud.net/}} lists 1'163 data sets from various domains for read-access.
The Linked Data Platform (LDP)\textsuperscript{\ref{foot:ldp}} specifies interaction with Read-Write Linked Data sources.
Next to projects such as Solid, these technologies offer read-write access to sensors and actuators on the Internet of Things (IoT), forming the Web of Things\textsuperscript{\ref{foot:wot}}. 
From such sensors and actuators, we can build applications such as integrated Cyber-Physical Systems, where sensors and actuators provide the interface to Virtual Reality systems (\cf the showcase in the evaluation, Section~\ref{example:ivision}).
Other non-RDF REST APIs provide access to weather reports\footnote{\url{http://openweathermap.org/}} or building management systems (\eg Project Haystack\footnote{\url{http://www.project-haystack.org/}}) and can be wrapped to support RDF.
Using such APIs, we can build applications such as integrated building automation systems (\cf the scenario of the synthetic benchmark in the evaluation, Section~\ref{sec:eval:empirical-smart-building}).

\todonote{Why hard?}

To use workflows in the environment of Read-Write Linked Data is difficult as the environment is fundamentally different from traditional environments where workflows are used.
\citeauthor*{DBLP:journals/fgcs/ElmrothHT10} argue that the properties of the environment determine the model of computation, which serves as the basis of a workflow language~\cite{DBLP:journals/fgcs/ElmrothHT10}.
Consequently, we have developed ASM4LD~\cite{DBLP:conf/www/KaferH18}, a model of computation for the environment of Read-Write Linked Data.
In this paper, we investigate an approach for a workflow language consisting of an ontology and operational semantics in ASM4LD.
In the investigation of the approach, the differences between traditional environments of workflow languages and the environment of Read-Write Linked Data (\ie RDF and REST) pose challenges:
\begin{description}
\item[Querying and reasoning under the open-world assumption]
Ontology \newline\mbox{languages} around RDF such as RDFS and OWL make the open-world assumption (OWA).
However, approaches from workflow management operate on relational databases, which make the closed-world assumption (CWA).
Closedness allows \eg to test if something holds for \emph{all} parts of a workflow.
\item[The absence of events in REST]
HTTP implements CRUD (the operations create, read, update, delete), but not the subscriptions to events.
However, approaches from workflow management use events as change notifications.
\end{description}
While both challenges could be mitigated by 
introducing assumptions (\eg negation-as-failure 
once we reach a certain completeness class~\cite{DBLP:conf/aaai/HarthS12})
or by extending the technologies (\eg implement events using Web Sockets\footnote{\url{http://www.ietf.org/rfc/rfc6455.txt}} or Linked Data Notifications~\cite{DBLP:conf/esws/CapadisliG0ASB17}), those mitigation strategies would restrict the generality of the approach, \ie we would have to exclude components that provide Linked Data, but do not share the assumptions or extensions of the mitigation strategy.

Previous works from Business Process Management, Semantic Web Services, Linked Data, and REST operate on a different model of computation or are complementary:
\cite{DBLP:conf/icsoc/PautassoW11,DBLP:conf/wsfm/HullDFGHHLMNSV10,DBLP:conf/icws/HallerCMOB05} assume event-based data processing, decision making based on process variables, and data residing in databases under the CWA, whereas our approach relies on integrated state information from the web under the OWA.
\cite{DBLP:conf/rest/VerborghSDCVW12,DBLP:conf/esws/ZaveriDWWVAKTJA17} provide descriptions to do automated composition or to assist developers.
Currently, we do not see elaborate and correct descriptions available at web scale, which hinders automated composition.
We see our approach, which allows for manual composition, as the first step towards automated composition.

\todonote{Overview of paper}
The paper is structured as follows:
In Section~\ref{sec:related-work}, we discuss related work.
In Section~\ref{sec:preliminaries}, we present the technologies on which we build our approach.
Next, we present our approach, which consists of two main contributions:
\begin{itemize}
\item An ontology to specify workflows and workflow instances modelled in OWL~LD\footnote{\url{http://semanticweb.org/OWLLD/}} (Section~\ref{sec:vocab}), which allows for querying and reasoning over workflows and workflow instances under the OWA.
The ontology is strongly related to the standard graphical workflow notation, BPMN, via the workflow patterns~\cite{DBLP:journals/dpd/AalstHKB03}.
\item An operational semantics for our workflow ontology.
We use ASM4LD, a model of computation for Read-Write Linked Data in the form of a condition-action rule language  
(Section~\ref{sec:semantics}), which
does not require event data and is directly executable.
We maintain workflow state in an LDP\footnote{\label{foot:ldp}\url{http://www.w3.org/TR/ldp/}} container.
\end{itemize}
Fast data processing thanks to OWL~LD and the executability of ASM4LD allow to directly apply our approach in practice. In the evaluation (Section~\ref{sec:evaluation}), we present a Virtual Reality showcase,
and a benchmark in an IoT setting.
We also show correctness and completeness of our approach.
We conclude in Section~\ref{sec:conclusion}.

\section{Related Work}
\label{sec:related-work}

We now survey related work grouped by field of research.

\begin{description}
\item[Workflow Management]
Previous work in the context of workflow languages and workflow management systems is based on event-condition-action (ECA) rules, whereas our approach is built for REST, and thus works without events.
This ECA rule-based approach has been used to give operational semantics to workflow languages~\cite{DBLP:conf/wsfm/HullDFGHHLMNSV10}, 
and to implement workflow management systems~\cite{DBLP:conf/dexa/CasatiCPP96}. 
Similar to the case handling paradigm~\cite{DBLP:journals/dke/AalstWG05}, we employ state machines for the activities of a workflow instance.

\item[Web Services]WS-*-based approaches assume arbitrary operations, whereas our approach works with REST resources, where the set of operations is constrained.
The BPM community has compared WS-*- and REST-based approaches~\cite{DBLP:conf/www/PautassoZL08,DBLP:journals/dss/MuehlenNS05}.
\citeauthor*{DBLP:conf/icsoc/PautassoW11} proposed extensions to BPEL such that \eg a BPEL process~\cite{DBLP:journals/dke/Pautasso09} can invoke REST services, and that REST resources representing processes push events~\cite{DBLP:conf/icsoc/PautassoW11}. 
While those extensions make isolated REST calls fit the Web Services processing model of process variable assignments, we propose a processing model based on integrated polled state.

\item[Semantic Web Services]
We can only present a selection of the large body of research conducted in the area of SWS.
Approaches like OWL-S
, WSMO 
and semantic approaches to scientific workflows like~\cite{DBLP:conf/iaai/GilRDMK07} are mainly concerned with service descriptions and corresponding reasoning for composition and provenance tracking.
In contrast to our work, which is based on REST, SWS build on Web Service technology for workflow execution, \eg 
the execution in the context of WSMO, WSMX~\cite{DBLP:conf/icws/HallerCMOB05}, is entirely event-based. 
European projects such as \enquote{Super} and \enquote{Adaptive Services Grid} build on WSMO.

\item[Ontologies for Workflows]
Similar to workflows in our ontology, processes in OWL-S are also tree-structured (see Section~\ref{sec:activities-workflows}) and use lists in RDF.
Unlike OWL-S, our ontology also covers workflow instances.
\citeauthor*{DBLP:conf/fois/RospocherGS14}~\cite{DBLP:conf/fois/RospocherGS14} and the project \enquote{Super} developed ontologies that describe process metamodels such as BPMN, BPEL, and EPC\footnote{\url{http://www.ip-super.org/content/view/129/136/}, \href{https://web.archive.org/web/20110726185128/http://www.ip-super.org/content/view/129/136/}{available in the Web Archive}}.
Their ontologies require more expressive (OWL) reasoning or do not allow for execution under the OWA.
\end{description}

\section{Preliminaries}
\label{sec:preliminaries}
We next introduce our environment, Read-Write Linked Data, and the model of computation, ASM4LD~\cite{DBLP:conf/www/KaferH18}, in which context we do querying and reasoning.

\subsubsection*{Read-Write Linked Data}

Linked Data is a collection of practices for data publishing on the web that advocates the use of web standards:
HTTP URIs\footnote{\url{http://www.ietf.org/rfc/rfc3986.txt}} should be used for identifying things.
HTTP GET\footnote{\url{http://www.ietf.org/rfc/rfc7230.txt}} requests to those URIs should be answered using descriptive data, \eg in RDF\footnote{\url{http://www.w3.org/TR/rdf11-concepts/}}.
Hyperlinks in the data should enable the discovery of more information\footnote{\url{https://www.w3.org/DesignIssues/LinkedData.html}}.
Read-Write Linked Data\footnote{\url{https://www.w3.org/DesignIssues/ReadWriteLinkedData.html}} introduces RESTful write access to Linked Data (later standardised in the LDP specification\textsuperscript{\ref{foot:ldp}}).
Hence, we can access the world's state using multiple HTTP GET requests and enact change using HTTP PUT, POST, DELETE requests.

In the paper, we denote RDF triples using binary predicates\footnote{We assume the URI prefix definitions of \url{http://prefix.cc/}
The empty prefix denotes \url{http://purl.org/wild/vocab\#}.
The base URIs be \nolinkurl{http://example.org/}.}, \eg
we write for the triple in Turtle notation \enquote{\nolinkurl{<\#wfm>} \nolinkurl{rdf:type} \nolinkurl{:WorkflowModel} \texttt{.}}:
\[
\mathtextit{rdf:type}(\mathtexttt{<\#wfm>},\mathtexttt{:WorkflowModel})
\]
We abbreviate a class assignment using a unary predicate with the class as predicate name, \eg $\mathtextit{:WorkflowModel}(\mathtexttt{<\#wfm>})$.
The term $\text{\renewcommand{\UrlFont}{\itshape}\nolinkurl{rdf:List}}(\dots)$ is a shortcut, similar to the RDF list shortcut with \texttt{()}  brackets in Turtle, and can be regarded as a procedure that
\begin{inparaenum}[(1)]
\item takes as argument list elements,
\item adds the corresponding RDF list triples, \ie with terms {\renewcommand{\UrlFont}{\itshape}\nolinkurl{rdf:first}}, {\renewcommand{\UrlFont}{\itshape}\nolinkurl{rdf:rest}}, and
\nolinkurl{rdf:nil}
, to the current data, and
\item returns the blank node for the RDF list's head.
\end{inparaenum}

\subsubsection*{ASM4LD, a Condition-Action Rule Language}
We use a monotonic production rule language to specify both forward-chaining reasoning on RDF data and interaction with Read-Write Linked Data resources~\cite{DBLP:conf/www/StadtmullerSHS13}.
A rule has a basic graph pattern\textsuperscript{\ref{foot:sparql}} query in the body.
We distinguish two types of rules:
\begin{inparaenum}[(1)]
\item derivation rules, which contain productions, \eg to implement reasoning, and
\item request rules, which specify an HTTP request in the rule head.
\end{inparaenum}
We assume safe rules and exclude existential variables in a rule head.
Rule programs consist of initial assertions and rules. 

As operational semantics for the rule language, we use ASM4LD, an Abstract State Machine-based~\cite{Gurevich:1995:EAL:233976.233979} model of computation for Read-Write Linked Data~\cite{DBLP:conf/www/KaferH18}.
In the following, we sketch the operational semantics, where data processing is done in repeated steps, subdivided into the following phases (\cf~\cite{DBLP:conf/www/KaferH18} for details):
\begin{enumerate}[(1)]
\item \label{list:ldf-step1}The working memory be empty.
\item \label{list:ldf-step2}Add the initial assertions to the working memory.
\item \label{list:ldf-step3}Evaluate on the working memory until the fixpoint:
\begin{enumerate}
\item Request rules that contain GET requests, making the requests and adding the data from the responses to the working memory.
\item Derivation rules, adding the produced data to the working memory.
\end{enumerate}
Using those GET requests and derivations, we acquire knowledge about the world's current state (from the responses to the GET requests) and reason on this knowledge (using the productions).
\item \label{list:ldf-step4} Evaluate all request rules with HTTP methods other than GET on the working memory and make the corresponding HTTP requests.
Using those requests, we enact changes on the world's state.
\end{enumerate}

To do polling, which is the prominent way 
to get information about changes in a RESTful environment, loop all of those phases (emptying the working memory each time step~\ref{list:ldf-step1} is reached).
Hypermedia-style link following (to discover new information) can be implemented using request rules, \eg in the example below.

We use the following rule syntax:
In the arguments of binary predicates for RDF, we allow for variables (printed in italics).
We print constants in typewriter font.
We connect rule head and body using $\rightarrow$.
The head of a request rule contains one HTTP request with the method as the function name, the target as the first argument, and the RDF payload as the second argument (if applicable).
E.\,g.\ consider the following rule to retrieve all elements $e$ of a given LDP container:
\begin{gather*}
\mathtextit{ldp:contains}(\mathtexttt{http://example.org/ldpc},e) 
\rightarrow\mathtextsc{get}(e)
\end{gather*}

\section{Activity, Workflow Model and Instance Ontology}
\label{sec:vocab}
To describe workflow models and instances as well as activities, we propose an ontology.
We developed the ontology, see Figure~\ref{fig:vocab}\footnote{The ontology can be accessed at \url{http://purl.org/wild/vocab}},
 with execution based on querying and reasoning under the OWA in mind.
In this section, we define activities, workflows, and instances using the workflow in Figure~\ref{fig:workflow} as example.
\begin{figure}[tb]
\begin{minipage}[b]{0.4\textwidth}
\includegraphics[scale=0.36]{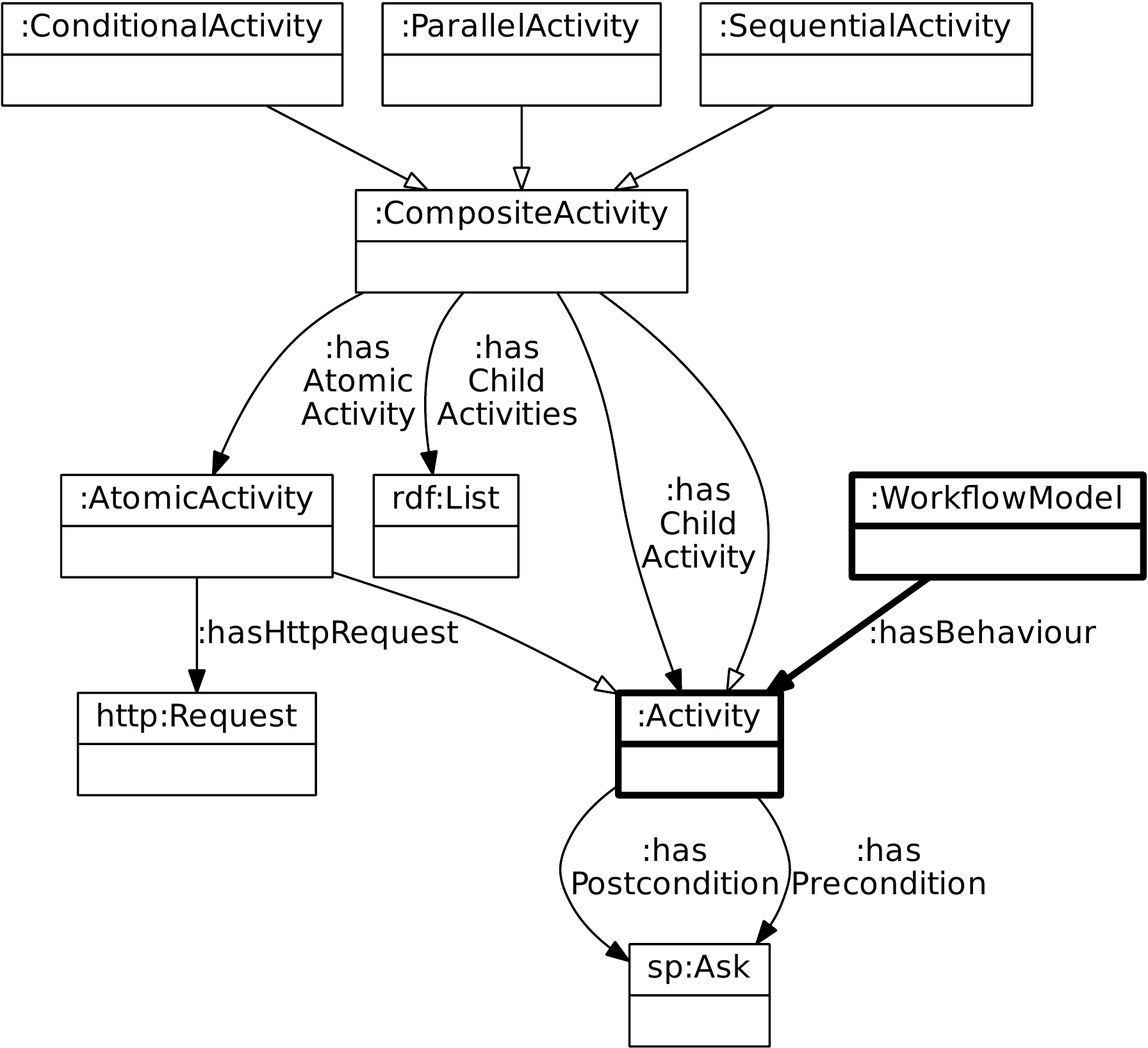}
\end{minipage}
\hfill
\begin{minipage}[b]{0.4\textwidth}
\includegraphics[scale=0.36]{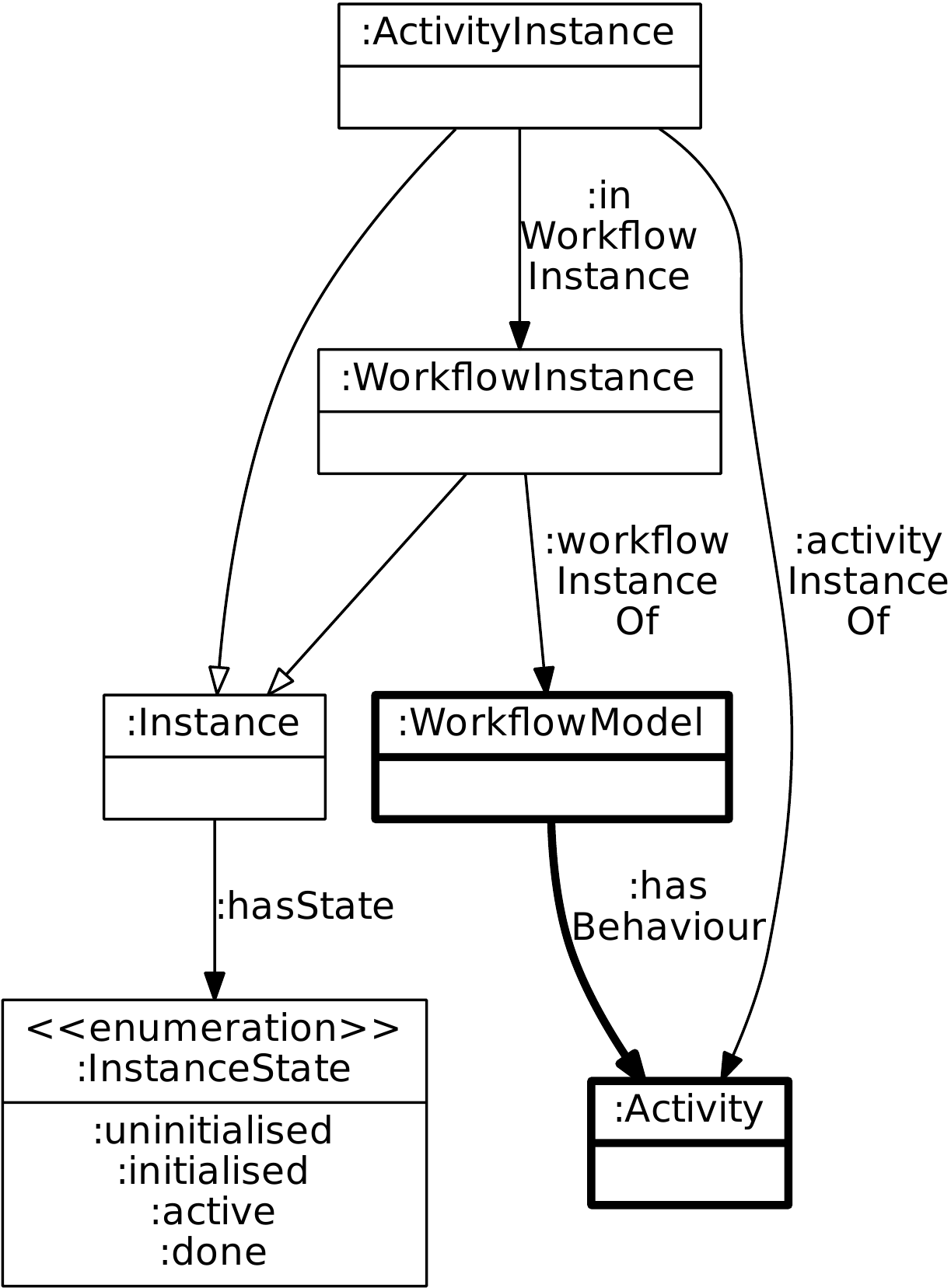}
\end{minipage}
\caption{\label{fig:vocab}The ontology to express workflow models and instances as UML Class Diagram.
Shared classes between the diagrams are depicted in bold.
We use the UML Class Diagram's class, inheritance, association, and enumeration to denote the RDFS ontology language's \nolinkurl{rdfs:Class}, \nolinkurl{rdfs:subClassOf}, \nolinkurl{rdf:Property} with \nolinkurl{rdfs:domain} and \nolinkurl{rdfs:range}, and instances.
}
\end{figure}
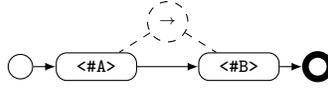
\begin{figure}[htb]
	\centering\footnotesize
	
	\begin{tikzpicture}[node distance=0.3cm]
	\tikzstyle{activity}	= [rectangle, draw, text centered, rounded corners, minimum height=1em, inner xsep=2ex, align=center]
	
	\node[dashed, circle, draw] (root) {\tiny$\rightarrow$};
	\node[activity] (ato1)[below left=of root]  {\scriptsize \verb|<#A>|};
	\node[activity] (ato2)[below right=of root] {\scriptsize \verb|<#B>|};
	\node[circle, draw, minimum height=1em] (start) [left=of ato1]  {} ;
	\node[circle, draw, minimum height=1em, line width=2.5pt] (end)   [right=of ato2] {} ;

	\draw[dashed] (root) -- (ato1);
	\draw[dashed] (root) -- (ato2);
    \draw[-latex] (ato1) -- (ato2);
    \draw[-latex] (start) -- (ato1);
    \draw[-latex] (ato2) -- (end);

	\end{tikzpicture}
	\caption{\label{fig:workflow}Workflow (solid: BPMN notation) with sequential activities (\texttt{<\#A>}, \texttt{<\#B>}). Dashed: the tree representation with the parent node marked as sequential.}
\end{figure}
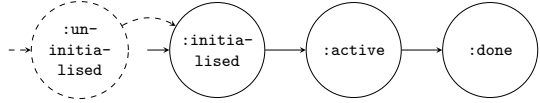
\begin{figure}[tb]
	\centering

\resizebox{!}{4em}{
	
	\begin{tikzpicture}[
	>=stealth,
	node distance=.25 and .75,
	every state/.style={minimum height=5.4em, font=\tt},
	]
	\node[initial,initial text=,state, align=center] (init)              {:initia-\\lised};
	\node[state]                       (active) [right=of init] {:active};
	\node[state]                       (done) [right=of active] {:done};

	\path[->] (init) edge[]           node[anchor=south] {} 
             (active);
	\path[->] (active) edge[]  node[anchor=south,align=center] {} 
                (done);

	\begin{scope}[dashed]
		\node[initial,initial text=,state, align=center] (un) [left=of init]             {:un-\\initia-\\lised};
		\path[->] (un) edge[bend left]  node[anchor=north] {}       (init);
	\end{scope}

	\end{tikzpicture}}
	
	\caption{\label{fig:PEandPrefixDFA}State machine for the instance resources for the workflow and activity instance resources.
	The dashed part only concerns workflow instance resources. 
	}
\end{figure}

\label{sec:activities-workflows}

\paragraph{Activities}
We regard an atomic activity as a basic unit of work.
We characterise an activity by a postcondition, which states what holds in the world's state after the activity has been executed.
We give the postcondition as a SPARQL ASK query\footnote{\label{foot:sparql}\url{http://www.w3.org/TR/sparql11-query/}}.
For the execution of an atomic activity, the activity description needs an HTTP request (\cf Figure~\ref{fig:vocab}).
The workflow model in Figure~\ref{fig:workflow} contains two activities (\texttt{<\#A>}, \texttt{<\#B>}), but omits postconditions and HTTP requests.

\paragraph{Workflow Models}
A workflow model is a set of activities put into a defined order.
As notation to describe workflow models, BPMN is a popular choice.
In the simplest form, the course of action (\ie control flow) in a BPMN workflow model is denoted using arrows that connect activities and gateways (\eg decisions and branches).
For instance, the middle arrow in the workflow model in Figure~\ref{fig:workflow} orders activities \texttt{<\#A>} and \texttt{<\#B>} sequentially.

In this paper, we assume a different representation of the control flow of a workflow model:
A tree structure, as investigated by \citeauthor*{DBLP:conf/bpm/VanhataloVK08}~\cite{DBLP:conf/bpm/VanhataloVK08}.
Tree-structured workflow languages include BPEL, a popular language to describe executable workflows.
In the tree, activities are leaf nodes.
The non-leaf nodes are typed, and the type determines the control flow of the children.
The connection between the tree-based (dashed) and the flow-based (solid) workflow representation is depicted in Figure~\ref{fig:workflow}.
Flow-based workflows can be losslessly translated to tree-structured workflows and vice versa~\cite{DBLP:conf/bpm/PolyvyanyyGD10}.
We use the tree structure, as checks for completion of workflow parts are easier in a tree.
Of the multitude of control flow features of different workflow languages, we support the most basic and common, which have been compiled to the \emph{basic workflow patterns}~\cite{DBLP:journals/dpd/AalstHKB03}.

We now show how to specify workflow models in RDF using Figure~\ref{fig:workflow}'s model:
\begin{gather*}
\mathtextit{:WorkflowModel}(\mathtexttt{<\#wfm>})\wedge\mathtextit{:SequentialActivity}(\mathtexttt{<\#root>})\\
\wedge\mathtextit{:AtomicActivity}(\mathtexttt{<\#A>})\wedge \mathtextit{:AtomicActivity}(\mathtexttt{<\#B>})\\
\wedge\mathtextit{:hasBehaviour}(\mathtexttt{<\#wfm>},\mathtexttt{<\#root>})\\
\wedge\mathtextit{:hasChildActivities}(\mathtexttt{<\#root>},\mathtextit{rdf:List}(\mathtexttt{<\#A>},\mathtexttt{<\#B>}))
\end{gather*}
As we assume tree-structured workflows, each workflow model (\verb|<#wfm>|) has a root activity (\verb|<#root>|).
If an activity is composite, \ie a control flow element, then the activity has an RDF list of child activities.
Here, \verb|<#root>| is sequential, with the child activities \verb|<#A>|, \verb|<#B>|.
The child activities could again be composite, thus forming a tree.
Leaves in the tree (here \verb|<#A>| and \verb|<#B>|) are atomic activities.
We require child activities to be given in an RDF list, which is explicitly terminated.
This termination closes the set of list elements and thus allows for executing workflows under the OWA, which \eg includes querying whether \emph{all} child activities of a parent activity are \texttt{:done}).
Yet, for the operational semantics we also need a direct connection between a parent activity and a child activity, which we derive from an RDF list using monotonic reasoning, here:
\begin{gather*}
\mathtextit{:hasChildActivity}(\mathtexttt{<\#root>},\mathtexttt{<\#A>})\wedge\mathtextit{:hasChildActivity}(\mathtexttt{<\#root>},\mathtexttt{<\#B>})
\end{gather*}

\paragraph{Instances}
Using workflow instances, we can run multiple copies of a workflow model.
A workflow instance consequently consists of instances of the model's activities.
We model the relation of the instances to their counterparts as shown in Figure~\ref{fig:vocab}.
During and after workflow monitoring/execution, the operational semantics maintain the states of instances in an LDP container.
At runtime, the instances' states evolve according to the state machine depicted in Figure~\ref{fig:PEandPrefixDFA} (terms from Figure~\ref{fig:vocab}).
Section~\ref{sec:semantics} is about the operationalisation of the evolution.

\section{Operational Semantics}
\label{sec:semantics}

In this section, we give operational semantics in rules\footnote{A corresponding Notation3 file can be found at \url{http://purl.org/wild/semantics}.} to our workflow language\footnote{In a production environment, access control to the instances' LDP container needs to be in place to keep third parties from interfering with the monitoring/execution.}.
Before we define the rules, we give an overview of what the rules do.

\subsection{Overview}
The rules fulfil the following purposes (the numbers are only to guide the reader):
\begin{enumerate}[I.]
\item \label{enum:retrieve-state}Retrieve state\footnote{A benefit of using Linked Data throughout is that we can access the workflow/activity instances' state and the world's state in a uniform manner.}
\begin{enumerate}[1)]
\item Retrieve the state of the writeable resources in the LDP container, which maintain the workflow/activity instances' state
\item Retrieve the relevant world state
\end{enumerate}
\item \label{enum:initialise}Initialise workflow instances if applicable
\begin{enumerate}[1)]
\item Set the root activity's instance \texttt{:active}
\item Set the workflow instance \texttt{:initialised}
\item Creating instance resources for all activities in the corresponding workflow model and set them \texttt{:initialised}
\end{enumerate}
\item \label{enum:finalise}Finalise workflow instances if their root node is \texttt{:done}
\item \label{enum:execute}Execute and observe \texttt{:active} activities
\begin{enumerate}[1)]
\item Execution: if an atomic activity turns \texttt{:active}, fire the HTTP request
\item If the postcondition of an \texttt{:active} activity is fulfilled, set it \texttt{:done}
\end{enumerate}
\item \label{enum:composite}Advance composite activities according to control flow, which includes: 
\begin{enumerate}[1)]
\item Set a composite activity's children \texttt{:active}
\item Advance between children
\item Finalise a composite activity
\end{enumerate}
\end{enumerate}

\subsection{Condition-Action Rules}
\label{sec:semantics:rules}
We next give the rules for the listed purposes.
To shorten the presentation, we factor out those rules that, for workflow execution, fire an activity's HTTP request if the activity becomes \texttt{:active}.
Those rules are not needed when monitoring.
The rules are of the form (the variable $\mathtextit{method}$ holds the request type):
\begin{gather*}
\mathtextit{AtomicActivity}(a)
\wedge\mathtextit{hasHttpRequest}(a,h)
\wedge\mathtextit{http:mthd}(h,method)\\
\wedge\mathtextit{http:requestURI}(h,u)
\wedge\dots
\rightarrow\mathtextsc{method}(u,\dots)
\end{gather*}

\subsubsection{\ref*{enum:retrieve-state}. Retrieve State}
The following rules specify the retrieval of data where the rule interpreter locally maintains state.
Analogously, the world state can get retrieved.
Either by explicitly stating URIs to be retrieved:
\begin{gather*}
\mathtexttt{true}
\rightarrow\mathtextsc{get}(\mathtexttt{http://example.org/ldpc})
\end{gather*}
Or by following links from data that is already known:
\begin{gather*}
\mathtextit{ldp:contains}(\mathtexttt{http://example.org/ldpc},e) 
\rightarrow\mathtextsc{get}(e)
\end{gather*}

\subsubsection{\ref*{enum:initialise}. Initialise Workflow Instances}
\label{sec:init}
If there is an uninitialised workflow instance (\eg injected by a third party using a \textsc{post} request into the polled LDP container), the following rules create corresponding resources for the activity instances and set the workflow instance initialised:
\begin{gather*}
\mathtextit{WorkflowInstance}(i)
\wedge\mathtextit{hasState}(i,\mathtexttt{:uninitialised})
\wedge\mathtextit{workflowInstanceOf}(i,m)\\
\wedge\mathtextit{hasBehaviour}(m,a)
\rightarrow\mathtextsc{post}(\mathtexttt{server:ldpc},\mathtextit{activityInstanceOf}(\mathtexttt{<\#it>},a)\\
\wedge\mathtextit{inWorkflowInstance}(\mathtexttt{<\#it>},i)\wedge\mathtextit{hasState}(\mathtexttt{<\#it>},\mathtexttt{:active}))
\end{gather*}
Also, the workflow instance is set initialised (analogously, we initialise instances for the activities in the workflow model):
\begin{gather*}
\mathtextit{WorkflowInstance}(i)
\wedge\mathtextit{hasState}(i,\mathtexttt{:uninitialised})
\wedge\mathtextit{workflowInstanceOf}(i,m)\\
\rightarrow\mathtextsc{put}(i,\mathtextit{WorkflowInstance}(i)\wedge\mathtextit{hasState}(i,\mathtexttt{:initialised})\\\wedge\mathtextit{workflowInstanceOf}(i,m))
\end{gather*}

\subsubsection{\ref*{enum:finalise}. Finalise Workflow Instances}
The done state of the root activity gets propagated to the workflow instance:
\begin{gather*}
\mathtextit{WorkflowInstance}(i)
\wedge\mathtextit{hasState}(i,\mathtexttt{:active})
\wedge\mathtextit{workflowInstanceOf}(i,m)\\
\wedge\mathtextit{hasBehaviour}(m,a)
\wedge\mathtextit{hasState}(m,\mathtexttt{:done})\\
\rightarrow
\mathtextsc{put}(i,\mathtextit{WorkflowInstance}(i)\wedge\mathtextit{hasState}(i,\mathtexttt{:done})\wedge\mathtextit{workflowInstanceOf}(i,m))
\end{gather*}

\subsubsection{\ref*{enum:execute}. Execute and Observe Atomic Activities}
In the following, we give a rule in its entirety, which marks an activity as done if its postcondition is fulfilled.
\begin{gather*}
\mathtextit{WorkflowInstance}(i)
\wedge\mathtextit{hasState}(i,\mathtexttt{:active})
\wedge\mathtextit{workflowInstanceOf}(i,m)\\
\wedge\mathtextit{hasDescendantActivity}(i,a)
\wedge\mathtextit{AtomicActivity}(a)
\wedge\mathtextit{hasPostcondition}(a,p)\\
\wedge\mathtextit{ActivityInstance}(j)
\wedge\mathtextit{activityInstanceOf}(j,a)
\wedge\mathtextit{hasState}(j,\mathtexttt{:active})\\
\wedge\mathtextit{sp:hasBooleanResult}(p,\mathtexttt{true})\\
\rightarrow
\mathtextsc{put}(j,\mathtextit{activityInstanceOf}(j,a)\wedge\mathtextit{inWorkflowInstance}(j,i)\wedge\mathtextit{hasState}(j,\mathtexttt{:done}))
\end{gather*}
To shorten the presentation of the rules, we introduce the following simplifications:
We assume that (1) we are talking about an active workflow instance, and (2) that the resource representing an instance coincides with its corresponding activity in the workflow model.
(3), the \textsc{put} requests in the text do not actually overwrite the whole resource representation but patch the resources by ceteris paribus overwriting the corresponding $\mathtextit{hasState}(\cdot,\cdot)$ triple.

\subsubsection{\ref{enum:composite}. Advance According to Control Flow}
\label{sec:semantics:rules:wfp}
In this subsection, we give the rules for advancing a workflow instance according to the basic workflow patterns~\cite{DBLP:journals/dpd/AalstHKB03}.

\paragraph{Workflow Pattern 1: Sequence}
\label{sec:semantics:rules:wfp:sequence}
If there is an active sequential activity with the first activity initialised, we set this first activity to active:
\begin{gather*}
\mathtextit{SequentialActivity}(s)
\wedge\mathtextit{hasState}(s,\mathtexttt{:active})
\wedge\mathtextit{hasChildActivities}(s,c)\\
\wedge\mathtextit{rdf:first}(c,a)
\wedge\mathtextit{hasState}(a,\mathtexttt{:initialised})
\rightarrow\mathtextsc{put}(a,\mathtextit{hasState}(a,\mathtexttt{:active}))
\end{gather*}
We advance between activities in a sequence using the following rule:
\begin{gather*}
\mathtextit{SequentialActivity}(s)
\wedge\mathtextit{hasState}(s,\mathtexttt{active})
\wedge\mathtextit{hasChildActivity}(s,c)\\
\wedge\mathtextit{hasState}(c,\mathtexttt{done})
\wedge\mathtextit{hasState}(n,\mathtexttt{initialised})\\
\wedge\mathtextit{rdf:first}(l,c)
\wedge\mathtextit{rdf:rest}(l,i)
\wedge\mathtextit{rdf:first}(i,n)
\rightarrow\mathtextsc{put}(n,\mathtextit{hasState}(n,\mathtexttt{active}))
\end{gather*}
If we have reached the end of the list of children of a sequence, we regard the sequence as done (the rule is an example of the exploitation of the explicit termination of the RDF list to address the OWA):
\begin{gather*}
\mathtextit{SequentialActivity}(s)
\wedge\mathtextit{hasState}(s,\mathtexttt{:active})
\wedge\mathtextit{hasChildActivity}(s,c)\\
\wedge\mathtextit{hasState}(c,\mathtexttt{:done})
\wedge\mathtextit{rdf:first}(l,c)
\wedge\mathtextit{rdf:rest}(l,\mathtexttt{rdf:nil})
\rightarrow\mathtextsc{put}(s,\mathtextit{hasState}(s,\mathtexttt{:done}))
\end{gather*}

\paragraph{Workflow Pattern 2: Parallel Split}
\label{sec:semantics:rules:wfp:parallel-split}
A parallel activity consists of several activities executed simultaneously.
If a parallel activity becomes active, all of its components are set to active:
\begin{gather*}
\mathtextit{ParallelActivity}(p)
\wedge\mathtextit{hasState}(p,\mathtexttt{active})
\wedge\mathtextit{hasChildActivity}(p,c)\\
\wedge\mathtextit{hasState}(c,\mathtexttt{:initialised})
\rightarrow\mathtextsc{put}(c,\mathtextit{hasState}(c,\mathtexttt{:active}))
\end{gather*}

\paragraph{Workflow Pattern 3: Synchronisation}
\label{sec:semantics:rules:wfp:synchronisation}
If all the components of a parallel activity are done, the whole parallel activity can be considered done.
To find out whether all components of a parallel are done, we have to mark instances the following way to deal with the RDF list.
First, we check whether the first element of the children of the parallel activity is done:
\begin{gather*}
\mathtextit{ParallelActivity}(p)
\wedge\mathtextit{hasState}(p,\mathtexttt{:active})
\wedge\mathtextit{hasChildActivities}(p,l)\\
\wedge\mathtextit{rdf:first}(l,c)
\wedge\mathtextit{hasState}(c,\mathtexttt{:done})
\rightarrow\mathtextit{hasState}(c,\mathtexttt{:doneFromListItemOne})
\end{gather*}
Then, starting from the first, we one by one check the activities in the list of child activities whether they are done.
If the check proceeded to the last list element, the whole parallel activity is done:
\begin{gather*}
\mathtextit{ParallelActivity}(p)
\wedge\mathtextit{hasState}(p,\mathtexttt{:active})
\wedge\mathtextit{hasChildActivity}(p,c)\\
\wedge\mathtextit{rdf:first}(l,c)
\wedge\mathtextit{rdf:rest}(l,\mathtexttt{rdf:nil})
\wedge\mathtextit{hasState}(c,\mathtexttt{:doneFromListItemOne})\\
\rightarrow\mathtextsc{put}(p,\mathtextit{hasState}(p,\mathtexttt{:done}))
\end{gather*}

\paragraph{Workflow Pattern 4: Exclusive Choice}
\label{sec:semantics:rules:wfp:choice}
The control flow element choice implements a choice between different alternatives, for which conditions are specified.
For the evaluation of the condition, we first have to check whether all child activities are in initialised state, similarly to the rules for Workflow Pattern 3:
\begin{gather*}
\mathtextit{ConditionalActivity}(a)
\wedge\mathtextit{hasState}(a,\mathtexttt{:active})
\wedge\mathtextit{hasChildActivities}(a,l)\\
\wedge\mathtextit{rdf:first}(l,c)
\wedge\mathtextit{hasState}(c,\mathtexttt{:initialised})\\
\rightarrow\mathtextit{hasState}(c,\mathtexttt{:initialisedFromListItemOne})\\
\mathtextit{ConditionalActivity}(a)
\wedge\mathtextit{hasState}(a,\mathtexttt{:active})
\wedge\mathtextit{hasChildActivities}(a,l)\\
\wedge\mathtextit{rdf:first}(l,c)
\wedge\mathtextit{hasState}(c,\mathtexttt{:initialisedFromListItemOne})\\
\wedge\mathtextit{rdf:rest}(l,m)
\wedge\mathtextit{rdf:first}(m,d)
\wedge\mathtextit{hasState}(d,\mathtexttt{:initialised})\\
\rightarrow\mathtextit{hasState}(d,\mathtexttt{:initialisedFromListItemOne})
\end{gather*}
If the check succeeded, we can evaluate the conditions and set an activity active:
\begin{gather*}
\mathtextit{ConditionalActivity}(a)
\wedge\mathtextit{hasState}(a,\mathtexttt{:active})
\wedge\mathtextit{hasChildActivitiy}(a,c)\\
\wedge\mathtextit{hasState}(c,\mathtexttt{:initialisedFromListItemOne})
\wedge\mathtextit{hasPrecondition}(c,p)\\
\wedge\mathtextit{rdf:first}(l,c)
\wedge\mathtextit{rdf:rest}(l,\mathtexttt{rdf:nil})
\wedge\mathtextit{sp:hasBooleanResult}(p,\mathtexttt{true})\\
\rightarrow\mathtextsc{put}(c,\mathtextit{hasState}(c,\mathtexttt{:active}))
\end{gather*}
We leave it to the modeller to make sure that the preconditions of the children of a conditional activity are mutually exclusive.

\paragraph{Workflow Pattern 5: Simple Merge}
\label{sec:semantics:rules:wfp:merge}
If one of the children of a conditional activity is done, the whole conditional activity is done:
\begin{gather*}
\mathtextit{ConditionalActivity}(a)
\wedge\mathtextit{hasState}(a,\mathtexttt{:active})
\wedge\mathtextit{hasChildActivitiy}(a,c)\\
\wedge\mathtextit{hasState}(c,\mathtexttt{:done})
\rightarrow\mathtextsc{put}(a,\mathtextit{hasState}(a,\mathtexttt{:done}))
\end{gather*}

\section{Evaluation}
\label{sec:evaluation}
First, we formally show the correctness of our approach to \emph{specifying} workflows by presenting the relationship of our operational semantics to the formal specification of the basic workflow patterns, which we support completely.
Second, to show the applicability of our approach in a real-world setting, we report on how we used the approach to do \emph{monitoring} of workflows for human-in-the-loop aircraft cockpit evaluation in Virtual Reality.
Third, we empirically evaluate our approach to \emph{executing} workflows in a Smart Building simulator.

\subsection{Mapping to Petri Nets}

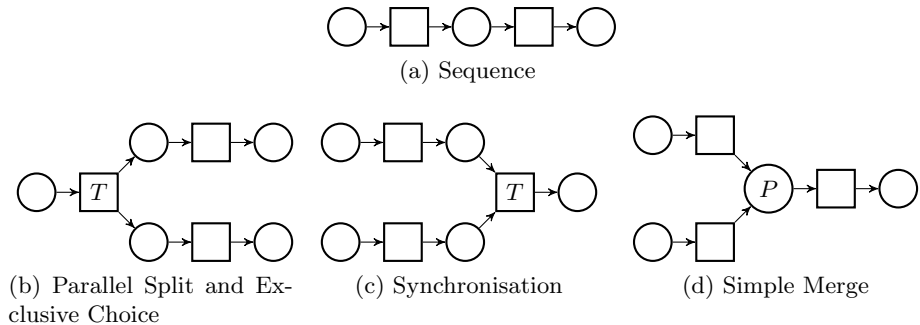
\begin{figure}[b]
\centering
\subfloat[\label{fig:wfp:sequence}Sequence]{
\begin{tikzpicture}[>=stealth',node distance=0.3cm]
\tikzstyle{place}      = [circle,   thick, draw, minimum size=5mm]
\tikzstyle{transition} = [rectangle,thick, draw, minimum size=5mm]

\node[place]      (a)                {} ;
\node[transition] (b) [right = of a] {} ;
\node[place]      (c) [right = of b] {} ;
\node[transition] (d) [right = of c] {} ;
\node[place]      (e) [right = of d] {} ;

\draw[->] (a) -- (b) ;
\draw[->] (b) -- (c) ;
\draw[->] (c) -- (d) ;
\draw[->] (d) -- (e) ;
\end{tikzpicture}}

\subfloat[\label{fig:wfp:split-choice}Parallel Split and Exclusive Choice]{
\begin{tikzpicture}[>=stealth',node distance=0.3cm]
\tikzstyle{place}      = [circle,   thick, draw, minimum size=5mm]
\tikzstyle{transition} = [rectangle,thick, draw, minimum size=5mm]

\node[place]      (a)                       {} ;
\node[transition] (b)  [right = of a]       {$T$} ;
\node[place]      (c1) [above right = of b] {} ;
\node[place]      (c2) [below right = of b] {} ;
\node[transition] (d1) [right = of c1]      {} ;
\node[transition] (d2) [right = of c2]      {} ;
\node[place]      (e1) [right = of d1]      {} ;
\node[place]      (e2) [right = of d2]      {} ;

\draw[->] (a)  -- (b)  ;
\draw[->] (b)  -- (c1) ;
\draw[->] (b)  -- (c2) ;
\draw[->] (c1) -- (d1) ;
\draw[->] (c2) -- (d2) ;
\draw[->] (d1) -- (e1) ;
\draw[->] (d2) -- (e2) ;
\end{tikzpicture}}
\hfill
\subfloat[\label{fig:wfp:synchronisation}Synchronisation]{
\begin{tikzpicture}[>=stealth',node distance=0.3cm]
\tikzstyle{place}      = [circle,   thick, draw, minimum size=5mm]
\tikzstyle{transition} = [rectangle,thick, draw, minimum size=5mm]

\node[place]      (a)                      {}  ;
\node[transition] (b)  [left = of a]       {$T$} ;
\node[place]      (c1) [above left= of b]  {}  ;
\node[place]      (c2) [below left= of b]  {}  ;
\node[transition] (d1) [left = of c1]      {}  ;
\node[transition] (d2) [left = of c2]      {}  ;
\node[place]      (e1) [left = of d1]      {}  ;
\node[place]      (e2) [left = of d2]      {}  ;

\draw[<-] (a)  -- (b)  ;
\draw[<-] (b)  -- (c1) ;
\draw[<-] (b)  -- (c2) ;
\draw[<-] (c1) -- (d1) ;
\draw[<-] (c2) -- (d2) ;
\draw[<-] (d1) -- (e1) ;of
\draw[<-] (d2) -- (e2) ;
\end{tikzpicture}
}
\hfill
\subfloat[\label{fig:wfp:simple-merge}Simple Merge]{
\begin{tikzpicture}[>=stealth',node distance=0.3cm]
\tikzstyle{place}      = [circle,   thick, draw, minimum size=5mm]
\tikzstyle{transition} = [rectangle,thick, draw, minimum size=5mm]

\node[place]      (a)                      {}  ;
\node[transition] (b)  [left = of a]       {}  ;
\node[place]      (c)  [left= of b]        {$P$} ;
\node[transition] (d1) [above left = of c] {}  ;
\node[transition] (d2) [below left = of c] {}  ;
\node[place]      (e1) [left = of d1]      {}  ;
\node[place]      (e2) [left = of d2]      {}  ;

\draw[<-] (a)  -- (b)  ;
\draw[<-] (b)  -- (c)  ;
\draw[<-] (c)  -- (d1) ;
\draw[<-] (c)  -- (d2) ;
\draw[<-] (d1) -- (e1) ;
\draw[<-] (d2) -- (e2) ;
\end{tikzpicture}
}
\caption{\label{fig:wfp:pnets}Petri Nets for the Basic Workflow Patterns.}
\end{figure}
\Citeauthor*{DBLP:journals/dpd/AalstHKB03} use Petri Nets to precisely specify the semantics of the basic workflow patterns (WFP)~\cite{DBLP:journals/dpd/AalstHKB03}.
We now show correctness by giving a mapping of our operational semantics to Petri Nets.
Similar to \textit{tokens} in a Petri Net that pass between \textit{transitions}, our operational semantics passes the \textit{active} state between \textit{activities} using rules (linking to the WFP rules from Section~\hyperref[sec:semantics:rules:wfp]{\ref*{sec:semantics:rules}.\ref*{enum:composite}}):
\begin{itemize}
\item The rule to advance between activities within a \texttt{:SequentialActivity} may only set an activity active if its preceding activity has terminated. 
In the Petri Net for the Sequence, a transition may only fire if the preceding transition has put a token into the preceding place, see Figure~\ref{fig:wfp:sequence} and the \hyperref[sec:semantics:rules:wfp:sequence]{WFP 1 rules}.
\item Only after the activity before a \texttt{:ParallelActivity} has terminated, the rule to advance in a parallel activity sets all child activities active.
In the Petri Net for the Parallel Split, all places following transition $T$ get a token iff transition $T$ has fired, see Figure~\ref{fig:wfp:split-choice} and the \hyperref[sec:semantics:rules:wfp:parallel-split]{WFP 2 rules}.

\item Only if all activities in a \texttt{:ParallelActivity} have terminated, the rules pass on the active state.
In the Petri Net for the Synchronisation, transition $T$ may only fire if there is a place with a token in all incoming arcs (\cf Figure~\ref{fig:wfp:synchronisation} and the \hyperref[sec:semantics:rules:wfp:synchronisation]{WFP 3 rules}).
\item In the \texttt{ConditionalActivity}, one child activity is chosen by the rule according to mutually exclusive conditions.
Similarly, exclusive conditions determine the continuation of the flow after transition $T$ in the Petri Net for the Exclusive Choice, see Figure~\ref{fig:wfp:split-choice} and the \hyperref[sec:semantics:rules:wfp:choice]{WFP 4 rules}.
\item If one child activity of a \texttt{:ConditionalActivity} switches from active to done, the control flow may proceed according to the rule.
Likewise, the transition following place $P$ in the Petri Net for the Simple Merge (Figure~\ref{fig:wfp:simple-merge}) may fire iff there is a token in $P$, \cf the \hyperref[sec:semantics:rules:wfp:merge]{WFP 5 rules}.
\end{itemize}

\subsection{Applicability: The Case of Virtual Aircraft Cockpit Design}
\label{example:ivision}

We successfully applied our approach together with industry in aircraft cockpit design~\cite{DBLP:conf/cspdata/KaferHM16}, where workflow monitoring is used to evaluate cockpit designs regarding Standard Operating Procedures.
The monitoring is traditionally done by Human Factors experts using 
stopwatches in physical cockpits.
We built an integrated Cyber-Physical System of Virtual Reality, flight simulation, sensors, and workflows to digitise the monitoring.
The challenge was to integrate the different system components on both the system interaction and the data level.
We built Linked Data interfaces to the components to cover the interaction integration, and used semantic reasoning to integrate the data.
Our approach allows monitoring workflows in this setting of Linked Data and semantic reasoning at runtime of the integrated system.
The integrated system also contains a UI to model workflows, which Human Factors experts evaluated as highly efficient.

\subsection{Empirical Evaluation using a Synthetic Benchmark}
\label{sec:eval:empirical-smart-building}
The scenario for our benchmark is from the IoT domain, where buildings are equipped with sensors and actuators from different vendors, which may be not interoperable.
This lack of interoperability has been identified by NIST as a major challenge for the building industry~\cite{gcr2004cost}.
\citeauthor*{DBLP:conf/sensys/BalajiBFGGHJKPA16} aim to raise interoperability in Building Management Systems (BMS) using Semantic Technologies:
Using the Brick ontology~\cite{DBLP:conf/sensys/BalajiBFGGHJKPA16}, people can model buildings and corresponding BMSs.
Read-Write Linked Data interfaces to a building's BMSs allow for executing building automation tasks.
Such tasks can include:
\begin{inparaenum}[(1)]
\item \label{enum:tasks:timecontrol}Control schemes based on time, sensor data and web data,
\item Automated supervision of cleaning personnel,
\item Presence simulation,
\item Evacuation support.
\end{inparaenum}
Those tasks go beyond simple rule-based automation tasks as typically found in home automation (\eg Eclipse SmartHome\footnote{\url{https://www.eclipse.org/smarthome/}}) and on the web (\eg IFTTT\footnote{\url{https://ifttt.com/}}), as the tasks require a notion of task instance state.
We therefore model the tasks as workflow and run them as workflow instances, which access the building management systems in an integrated fashion using Read-Write Linked Data interfaces.

The environment for our benchmark is a Linked Data representation of building 3 of IBM Research Dublin:
Based on a static description of building 3\footnote{\label{foot:IBM_B3}\url{https://github.com/BuildSysUniformMetadata/GroundTruth/blob/2e48662/building_instances/IBM_B3.ttl}} using the Brick ontology~\cite{DBLP:conf/sensys/BalajiBFGGHJKPA16}, which covers the building’s parts (\eg rooms) and the parts of the building’s systems (\eg lights and switches), we built a Linked Data representation as follows.
We subdivided the description into one-hop RDF graphs around each URI from the building and provide each graph for dereferencing at the corresponding URI.
No data is lost in the subdivision, as there are no blank nodes in the description.
To add state information to the systems, we add writeable SSN\footnote{\url{https://www.w3.org/TR/vocab-ssn/}} properties to the Linked Data interface.
To evaluate at different scales, we can run multiple copies of the building.

The workload for our benchmark is the control flow of the five representative workflow models proposed by \citeauthor*{DBLP:conf/caise/FermeSIPL17}~\cite{DBLP:conf/caise/FermeSIPL17} for evaluating workflow engines, determined by clustering workflows from literature, the web, and industry. 
We interpreted the five workflow models using the four automation tasks presented above:
Task~\ref{enum:tasks:timecontrol} corresponds to the first two workflow models; the subsequent tasks to the subsequent workflow models.
We distinguish two types of activities in the tasks: activities that are mere checks, \ie have only a postcondition (\eg an hour of the day to build time-based control), and tasks that enact change (\eg turn on a light), where we attach an HTTP request.
We assigned the types to the workflows' activities and made sure, for repeatability, that the postconditions are always fulfilled and that the requests do not interfere with the workflow.

\todo{What's the tense? Present or past? Use consistently.}
The set-up for our evaluation using the benchmark contains a server with a 32-core Intel Xeon E5-2670 CPU and 256\,GB of RAM running Debian Jessie.
We use the server to run the buildings and the workflow management.
For the workflow management, we deploy the operational semantics and required OWL~LD reasoning on Linked Data-Fu 0.9.12\footnote{\url{http://linked-data-fu.github.io/}}.
We add rules for inverse properties as presented in the examples of the Brick ontology.
The buildings and the workflow state are maintained in individual Eclipse Jetty servers running LDBBC 0.0.6\footnote{\url{http://github.com/kaefer3000/ldbbc}} LDP implementations. 
We add workflow instances each 0.2\,s after a warm-up time of 20\,s.
The workflow models and more information can be found online\footnote{\url{http://people.aifb.kit.edu/co1683/2018/iswc-wild/}}.

The results of our evaluation can be found in Table~\ref{tab:empirical-results}.
\begin{table}[tb]
\centering
\caption{\label{tab:empirical-results}Average runtime [s] for workflows W$n$ in different numbers of buildings.}
\begin{tabular}{rrrrrr}
\toprule
& W1 & W2 & W3 & W4 & W5 \\
\midrule
1 Building\hspace*{1ex} & 2& 2& 6& 12& 18\\
10 Buildings & 8& 9&26& 61& 75 \\
20 Buildings &12&13&38& 80&109 \\
50 Buildings &19&21&61&156&218 \\
\bottomrule
\end{tabular}
\end{table}
Varying the number of activities (W1-W5), and varying the number of devices (proportional to buildings), we observe linear behaviour.
The linear behaviour stems from the number of requests to be made, which depends on the number of activities and workflow instances.
With no data and reasoning results reusable between buildings, there is no benefit in running the workflows for all buildings on one engine.
Instead, we could run one engine per building, thus echoing the decentralisation of data.

\section{Conclusion}
\label{sec:conclusion}

We presented an approach to specify, monitor, and execute applications that build on distributed data and functionality provided as Read-Write Linked Data.
We use workflows to specify applications, and thus defined a workflow ontology and corresponding operational semantics to monitor and execute workflows.
We aligned our approach to the basic workflow patterns, reported on an application in Virtual Reality, and evaluated using a benchmark in an IoT scenario.

The assumptions of the environment of Read-Write Linked Data (\ie RDF and REST) present peculiar challenges for a workflow system:
We work under the open-world assumption and without notifications.
Our approach addresses the challenges without adding assumptions to the architecture of the environment, but by modelling a closed world where necessary and by using polling.

We believe that our approach, which brings workflows in a language that is closely related to the popular BPMN notation to Read-Write Linked Data, enables non-experts to engage in the development of applications for Read-Write Linked Data that can be verified, validated, and executed.

\section*{Acknowledgements}
We acknowledge helpful feedback by Rik Eshuis
and Philip Hake on an earlier draft of the paper.
This work is partially supported by the EU's FP7 under GA~605550, i\nobreakdash-VISION, and by the German BMBF under FKZ~01IS12051, AFAP.

\section*{References}
\printbibliography[heading=none]
\end{document}